\documentclass[lettersize,journal]{IEEEtran}
\usepackage{amsmath,amsfonts}
\usepackage{algorithmic}
\usepackage{algorithm}
\usepackage{array}
\usepackage[caption=false,font=normalsize,labelfont=sf,textfont=sf]{subfig}
\usepackage{textcomp}
\usepackage{stfloats}
\usepackage{url}
\usepackage{verbatim}
\usepackage{graphicx}
\usepackage{cite}
\usepackage{xspace}
\usepackage[table]{xcolor}
\colorlet{gainbg}{green!10}
\colorlet{dropbg}{red!8}
\usepackage{multirow}
\usepackage{booktabs}
\begin{document}
\bstctlcite{BSTcontrol}
\newcommand{\modelname}{\textsc{MoE-Prism}\xspace}

\title{\modelname: Model and System Support for Request-Level Compute Elasticity in MoE Serving}

\author{Xinfeng Xia, Xiaofeng Hou~\IEEEmembership{Member,~IEEE}, Jiacheng Liu~\IEEEmembership{Member,~IEEE} , Wenfeng Wang, Mingxuan Zhang, Peng Tang, Chao Li~\IEEEmembership{Senior Member,~IEEE}, Minyi Guo~\IEEEmembership{Fellow,~IEEE}

\thanks{This work has been submitted to the IEEE for possible publication. Copyright may be transferred without notice, after which this version may no longer be accessible.}

\IEEEcompsocitemizethanks{
\IEEEcompsocthanksitem X. Xia, X. Hou, W. Wang, M. Zhang, P. Tang, C. Li are with the School of Computer Science, Shanghai Jiao Tong University.

J. Liu is with the Hong Kong University of Science and Technology.

M. Guo is with the School of Computer Science, Shanghai Jiao Tong University and the State Key Laboratory of Public Big Data, Guizhou University, Guiyang, China.

}}

\maketitle

\begin{abstract}

Mixture-of-Experts (MoE) scales model capacity through sparse activation, and is becoming an important architecture for large language models (LLMs). However, existing MoE serving systems typically execute all requests under a fixed routing configuration, limiting their ability to exploit heterogeneous computation requirements across requests. Routing top-$k$, which determines the number of routed experts activated per token, directly controls routed-expert computation and provides a natural mechanism for request-level compute elasticity. Realizing this capability, however, requires finer-grained routing units and efficient runtime execution for heterogeneous routing budgets.
We present \modelname, a model and system support framework for request-level compute elasticity in MoE serving. \modelname decomposes monolithic experts into fine-grained sub-experts to expose denser routing operating points and provides a $k$-aware serving runtime that effectively serves heterogeneous routing budgets under both throughput-oriented and latency-sensitive workloads.
We implement \modelname on top of vLLM and evaluate it on three representative MoE models. \modelname expands the number of available routing operating points by $4\times$, improves offline inference throughput by up to 33.9\%, and reduces online serving TTFT under heterogeneous workloads. These results demonstrate practical elastic MoE serving with request-level routing targets.

\end{abstract}

\section{Introduction}
\IEEEPARstart{M}{ixture-of-Experts} (MoE) architectures have become a prominent design choice for frontier large language models (LLMs), scaling model capacity without a proportional increase in computation per token~\cite{jiang2024mixtral, yang2025qwen3, liu2024deepseek, team2025kimi}.
As these models move into production, serving systems must handle requests with heterogeneous quality requirements. For example, casual conversations may be adequately served with a small computation budget, whereas professional research tasks may require substantially more computation to meet their quality targets.

We call the ability to serve different requests with different amounts of computation according to their serving requirements \emph{request-level compute elasticity}. Existing LLM routing and cascading approaches realize this idea by selecting among models with different cost--quality profiles or escalating to stronger models when needed~\cite{chen2023frugalgpt,ong2025routellm}. However, they achieve elasticity through multiple independently deployed models, which introduces additional memory and deployment costs. This motivates exposing multiple operating points within a single resident model.

Doing so requires an internal computation knob within a single model. One option is to shorten the input through context compression or truncation~\cite{jiang2023llmlingua,jiang2024longllmlingua}. This can reduce inference cost, but may discard information essential to the request, particularly in long-context tasks. Another option is to reduce the executed depth through early-exit and dynamic-depth methods~\cite{bae2025mix,schuster2022confident,elhoushi2024layerskip}. These methods change computation at whole-layer granularity and often require specialized training or architectural support.

We observe that MoE routing top-$k$, i.e., the number of routed experts activated per token, provides a natural mechanism for realizing request-level compute elasticity.
Varying $k$ directly adjusts routed-expert computation, which scales approximately linearly with $k$, while providing different computation--capability trade-offs within a single deployed model.
Most pretrained MoE models nevertheless retain their training-time routing budget at inference. Recent work has relaxed this constraint by allowing tokens to select different expert counts~\cite{guo2024dynamic}, training models to operate across multiple inference-time expert counts~\cite{gu2025elastic}, or dynamically varying expert precision~\cite{wang2025d2moe}. These studies establish the model-side feasibility of varying top-$k$. However, their operating points remain at the native whole-expert granularity, and do not address how to execute these request-level heterogeneity effectively.

\begin{figure}[t]
    \centering
    \includegraphics[width=1\linewidth]{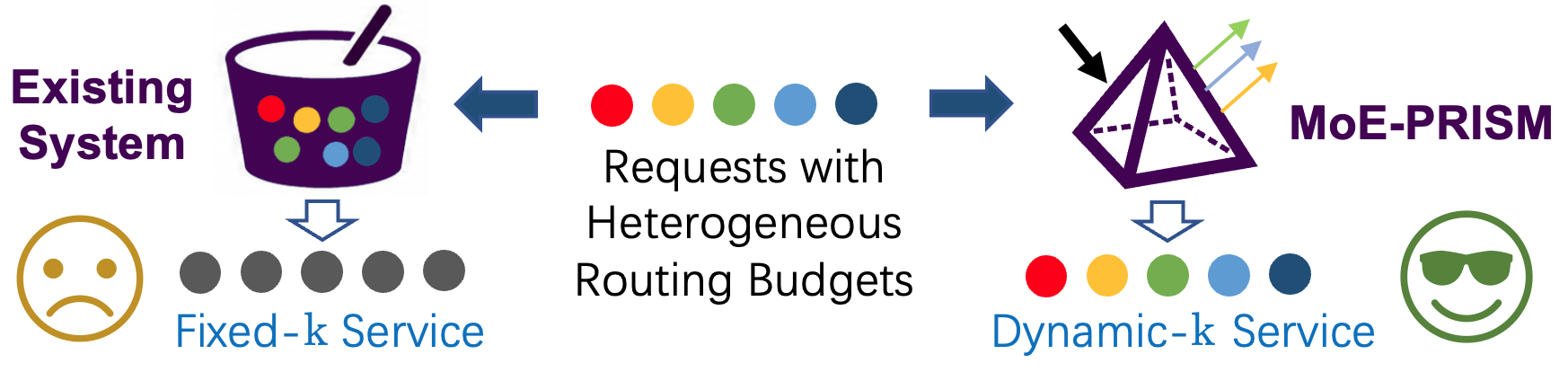}
    \caption{Fixed-$k$ serving homogenizes heterogeneous request budgets and wastes computation, whereas \modelname uses efficient dynamic-$k$ execution to reduce computation while preserving request-level budgets.}
    \label{fig:intro-motivation}
\end{figure}

To our knowledge, no existing MoE serving system jointly exposes externally specified, per-request top-$k$ targets and executes them efficiently within a single serving instance. 
These target $k$ may be supplied by a user, service tier, or upstream policy.
As illustrated in Fig.~\ref{fig:intro-motivation}, fixed-$k$ serving collapses these budgets to a single operating point, whereas our goal is to preserve them within the same serving instance and translate the resulting compute savings into end-to-end latency or throughput gains. 
Realizing this interface requires support at both the model and system levels.

\textit{First, pretrained MoE models expose only coarse routing operating points.}
Recent MoE models scale capacity by expanding the expert pool while activating only a small subset for each token~\cite{jiang2024mixtral, yang2025qwen3, liu2024deepseek}.
For example, \texttt{Kimi K2}~\cite{team2025kimi} scales to roughly 1T parameters while routing each token to only eight routed experts, in addition to one shared expert.
Because each expert is treated as an indivisible routing unit, reducing $k$ by one removes an entire expert and a substantial fraction of routed-expert computation, particularly at smaller budgets.
A \emph{routing operating point} represents a supported routing budget together with its corresponding computation cost and model quality, describing one point in the computation--quality trade-off space.
Adjacent routing operating points can therefore differ sharply in both computation cost and model quality, limiting request-level compute elasticity.
Our measurements further show that activations within an expert are concentrated in token-dependent subsets of neurons, yet the native routing interface cannot exploit this internal structure to expose intermediate operating points.

\textit{Second, existing serving systems are not designed to preserve heterogeneous routing budgets efficiently.}
MoE runtimes optimize shared execution around a uniform routing configuration, whereas request-level compute elasticity introduces different routing budgets into the same serving pipeline. Without dedicated support, homogenizing these budgets forfeits the expected computation savings, while preserving them can undermine batching and execution efficiency. Therefore, practical request-level compute elasticity requires runtime mechanisms that retain per-request computation savings without compromising high-throughput serving.

In this paper, we present \modelname, a model and system support framework for request-level compute elasticity in MoE serving.
\modelname enables a single MoE serving instance to efficiently execute requests with heterogeneous routing budgets.
On the model side, the \emph{Model Refactoring Engine} profiles intermediate neuron activations offline, partitions each original expert into smaller sub-experts whose neurons tend to co-activate, and reconstructs the routing interface over the expanded expert set. This refactoring reuses the pretrained expert parameters and enables a single refactored model to expose a denser spectrum of routing operating points. It supports a training-free deployment path and, optionally, lightweight gate-only fine-tuning, without requiring end-to-end retraining.
On the system side, the \emph{$k$-Aware Serving Scheduler} groups and executes requests according to their routing budgets, reducing the excess computation caused when heterogeneous requests are mixed under a larger routing configuration.
Together, these components introduce finer-grained operating points and preserve their computational benefits under heterogeneous serving workloads in a real-world MoE serving stack.

Overall, we make the following contributions:
\begin{itemize}
\item We introduce a model refactoring method that decomposes monolithic experts into finer-grained sub-experts and reconstructs their gating network, enabling a single refactored model to expose denser request-level routing operating points. The method supports a training-free deployment path and optional gate-only fine-tuning, without requiring end-to-end retraining.
\item We extend \texttt{vLLM} with a $k$-aware serving runtime that combines budget-aware scheduling with efficient dynamic-$k$ execution, preserving the computation savings of heterogeneous routing budgets during batched serving.
\item We integrate \modelname into \texttt{vLLM} and evaluate it on three representative MoE models. \modelname improves offline throughput by up to 33.9\% and reduces online Time-to-First-Token (TTFT) under heterogeneous workloads.
\end{itemize}

\section{Motivation and Challenges}
\label{sec:bg-moti}

\begin{figure}[t]
    \centering
    \includegraphics[width=1\linewidth]{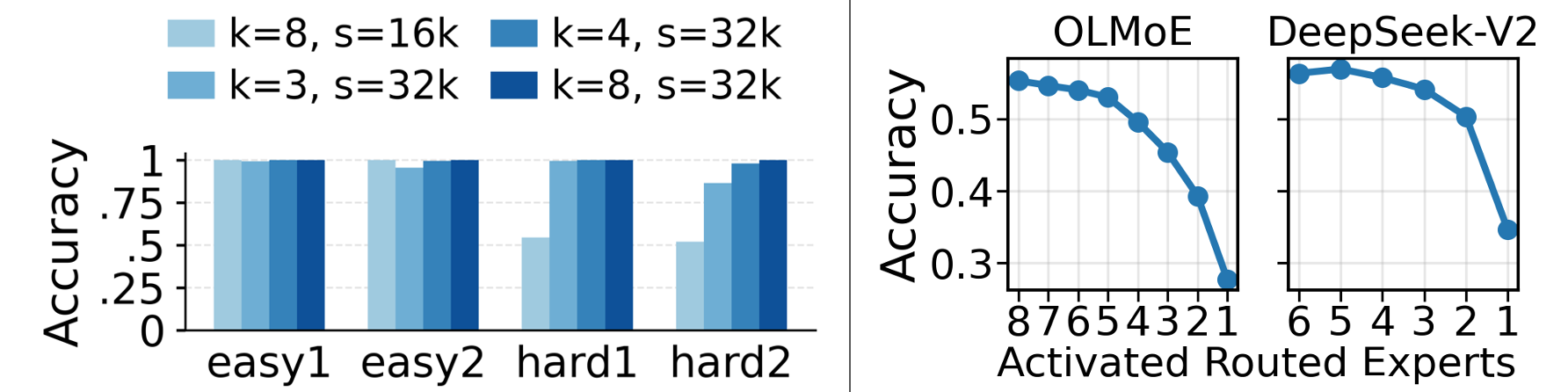}
    \caption{
    Left: Qwen3-30B-A3B accuracy across routed-expert counts $k$ and inference input limits $s$. Easy and hard denote short and long test samples. 1 and 2 denote the \texttt{niah\_single} and \texttt{niah\_multikey} tasks. \\ Right: ARC-Challenge accuracy under different routing budgets.}
    \label{fig:bg-acc}
\end{figure}

\subsection{Routing Budget Directly Controls MoE Workload}
\label{sec:moti-workload}

In MoE layer execution, each processed token is dispatched to $k$ routed experts. For an input that reaches the inference limit $s$, this produces $s \times k$ token--expert assignments. With the token count and model dimensions fixed, this quantity linearly scales the logical input volume of the routed-expert GEMMs. Thus, $k$ provides direct control over routed-expert computation, although the realized latency and throughput benefits depend on batching and hardware efficiency. The remaining question is whether pretrained models retain useful quality below their native routing budget.

Figure~\ref{fig:bg-acc} (left) examines this trade-off on \texttt{Qwen3-30B-A3B} using two RULER retrieval tasks~\cite{hsieh2024ruler}: \texttt{niah\_single}, which retrieves one hidden value, and \texttt{niah\_multikey}, which must recover multiple values. Here, easy and hard describe the original test-sample lengths (16k for easy and 32k for hard), whereas $s$ denotes the model-side inference input limit. Lowering $s$ below the original sample length truncates the input and can sharply degrade accuracy once relevant evidence falls outside the retained window, whereas preserving the full input and lowering $k$ retains substantially higher task accuracy under a comparable routed-expert workload. Specifically, for the hard tasks, reducing $k$ from 8 to 4 halves the number of token--expert assignments while preserving 99.5\% and 98.0\% accuracy on the two tasks, respectively. By contrast, setting $s=16k$ for the same examples while retaining $k=8$ reduces accuracy to 54.5\% and 52.0\%. These results illustrate $k$ as a meaningful computation--quality control dimension distinct from context truncation.

\subsection{Expert-Level Routing Provides Sparse and Coarse Control}
\label{sec:moti-coarse}

Native MoE routing adjusts computation only by adding or removing whole routed experts. Throughout this paper, $k$ denotes the number of selected routed experts, excluding any shared experts. Since pretrained MoE models typically activate only a small number of routed experts per token, scaling down from the native activation count offers only a few discrete $k$ settings. These settings can also be too coarse at low $k$, where removing a single expert may cause a disproportionate loss in accuracy.

Figure~\ref{fig:bg-acc} (right) illustrates this limitation on ARC-Challenge across two MoE models. OLMoE-1B-7B degrades gradually from $k=8$ to $k=5$, but its accuracy drops much more sharply at lower $k$. DeepSeek-V2-Lite remains nearly stable from $k=6$ to $k=2$, yet drops sharply when $k$ is reduced from $2$ to $1$. These results reveal the lack of granularity: adjacent routing points fail to provide a smooth computation--quality trade-off; particularly in the low-\(k\) regime, a single-step reduction in \(k\) precipitates a disproportionately severe degradation in model accuracy.

\subsection{Activations Are Concentrated Within Expert Neurons}
\label{sec:moti-concentration}

\begin{figure}[t]
    \centering
    \includegraphics[width=1\linewidth]{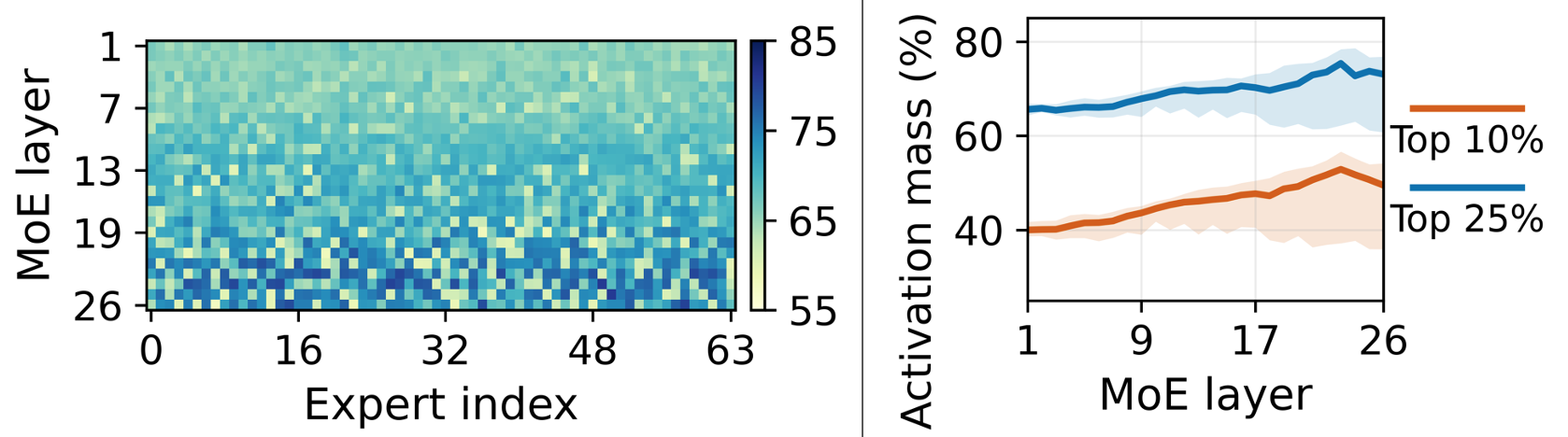}
    \caption{Left: Activation concentration across all 1,664 layer-expert pairs in DeepSeek-V2-Lite. Each cell reports the mean mass carried by a routed token's top 25\% intermediate neurons. \\ Right: Activation concentration across model depth of DeepSeek-V2-Lite. Lines show layer-wise medians and shaded bands show the 10th--90th percentiles across experts.}
    \label{fig:cal-proj}
\end{figure}

The sparse native control surface raises a natural question: is a full expert the appropriate atomic routing unit? Each MoE expert contains an FFN that expands the token representation into a higher-dimensional intermediate activation vector. We refer to each element of this vector as an \emph{intermediate neuron}. To examine how activation is distributed among these neurons, we profile all 1,664 layer--expert pairs in the 26 MoE layers of \texttt{DeepSeek-V2-Lite}, using 128K tokens in \texttt{WikiText-2} training split for calibration. For each token routed to an expert, we independently rank the intermediate neurons by their absolute activation values.
The top 10\% (or top 25\%) denotes the corresponding fraction of neurons with the largest absolute activations for that token.
We compute the fraction of the token's total $L_1$ activation mass carried by these neurons. Each heatmap cell averages this fraction across tokens routed to one layer--expert pair. The reported median then aggregates across the 1,664 pairs. Normalizing by total activation mass permits comparisons across layers and experts with different activation scales.
Activation concentration is widespread and consistent (Figure~\ref{fig:cal-proj}). Across the 1,664 layer--expert pairs, the top 25\% of neurons carry 67.9\% of the activation mass at the median. Breaking this statistic down by model depth, the top-10\% and top-25\% medians remain high across all 26 layers.

Different tokens produce different activation patterns within the same expert: neurons that are highly activated for one token may be unimportant for another. A token-agnostic split, such as a contiguous or uniformly balanced partition, would ignore which neurons tend to become important together and could scatter a token's activation mass across many units. This creates a model-side challenge: how can we derive a fixed partition from token-dependent activations, so that neurons with correlated activation profiles across tokens are clustered into the same routable unit? Such a partition would concentrate each token's activation mass into fewer fine-grained units and enable routing budgets between adjacent whole-expert settings.

\subsection{Heterogeneous Routing Budgets Challenge Batched Execution}
\label{sec:moti-batching}

\begin{figure}[t]
    \centering
    \includegraphics[width=0.8\linewidth]{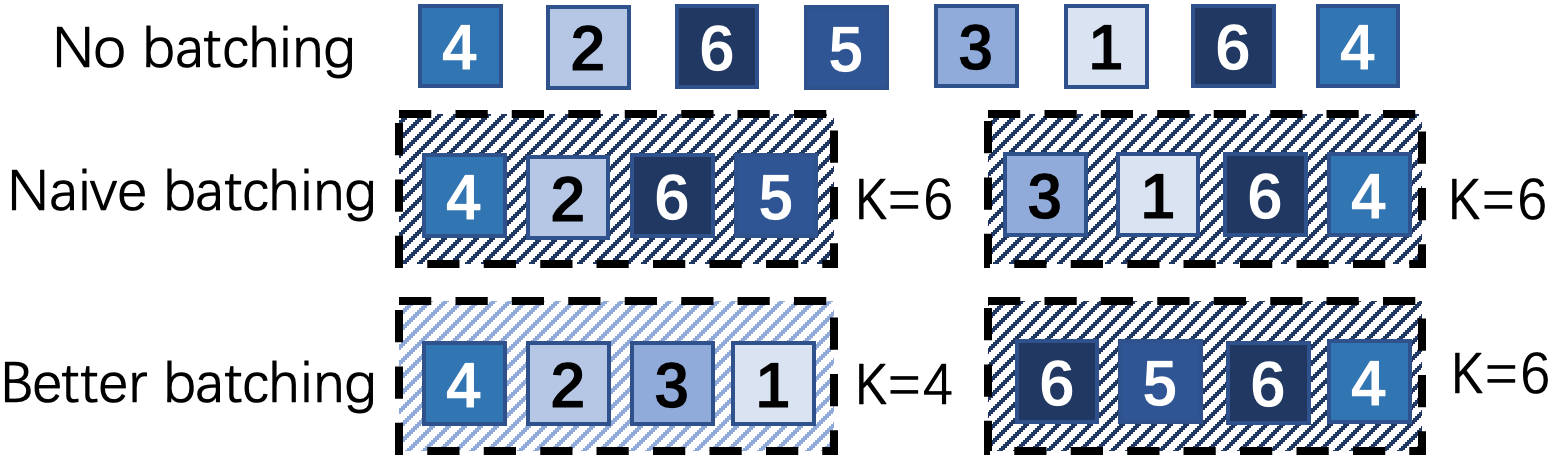}
    \caption{Mixed-$k$ batching strategies. A batch-wide maximum over-serves low-$k$ requests, while grouping similar budgets reduces this waste.}
    \label{fig:sche-bg-right}
\end{figure}

Lowering $k$ reduces the routed-expert workload of an individual request, but conventional MoE kernels using grouped-GEMM typically assume a uniform top-$k$ across all tokens in an invocation. Serving engines such as \texttt{vLLM}~\cite{kwon2023efficient} use continuous batching to combine tokens from many requests into shared execution batches. When requests in the same batch carry different routing budgets, these two properties create a trade-off between over-serving and batch fragmentation (Figure~\ref{fig:sche-bg-right}).

Consider a batch containing requests with budgets $k\in\{2,4,5,6\}$. Executing every token at the batch-wide maximum, $k=6$, preserves one large execution batch but performs unnecessary routed-expert computation for the other requests, partially or entirely erasing their intended computation savings. Alternatively, partitioning requests by budget avoids this over-serving but breaks one large batch into several smaller invocations. GEMM operations on smaller token groups are less efficient, while each partition incurs additional invocation and kernel-launch overheads, further reducing overall execution efficiency.
Therefore, an efficient top-$k$ control requires runtime scheduling policies that preserve routing heterogeneity instead of collapsing mixed requests into a single higher-$k$ strategy.

\subsection{Design Requirements}

Taken together, these measurements reveal both an opportunity and a systems gap. At a fixed input length, lowering $k$ directly reduces routed-expert work while retaining high accuracy over a useful range (\S\ref{sec:moti-workload}). However, native whole-expert routing exposes too few computation--quality operating points (\S\ref{sec:moti-coarse}), even though activation concentration below the expert level suggests that finer routing units are possible (\S\ref{sec:moti-concentration}). Moreover, heterogeneous batching can erode the computation savings provided by per-request budgets (\S\ref{sec:moti-batching}). Making request-level $k$ practical therefore imposes one requirement on the model and one on the serving system.

\textbf{R1: A denser and usable routing space.} The model must expose useful computation--quality operating points between adjacent whole-expert budgets. Because neuron importance varies across tokens, the resulting fixed routing units should capture recurring neuron co-activation patterns across inputs rather than rely on token-agnostic partitions. The transformation should reuse the pretrained expert parameters and avoid expensive end-to-end retraining.

\textbf{R2: Budget preservation under heterogeneous batching.} The runtime must limit the over-serving caused by batch-wide maximum budgets without excessively fragmenting execution into small, inefficient groups. It must therefore balance fidelity to per-request routing budgets against the batching efficiency required for high-throughput serving. Operationally, to control the CUDA-graph memory overhead introduced by the expanded routing space, we additionally share captured attention segments across budgets and specialize only the MoE segments.

\modelname satisfies R1 through its \emph{Model Refactoring Engine} (\S\ref{sec:model-side}), which constructs finer-grained routing units and reconstructs their gating interface. Its \emph{$k$-Aware Serving Runtime} satisfies R2 through a \emph{$k$-Aware Runtime Scheduler} (\S\ref{sec:system-implementation}) and a \emph{Segmented Graph Executor} (\S\ref{sec:executor}).

\section{The \modelname Design}

\subsection{Overview of \modelname}
\begin{figure}[t]
    \centering
    \includegraphics[width=1\linewidth]{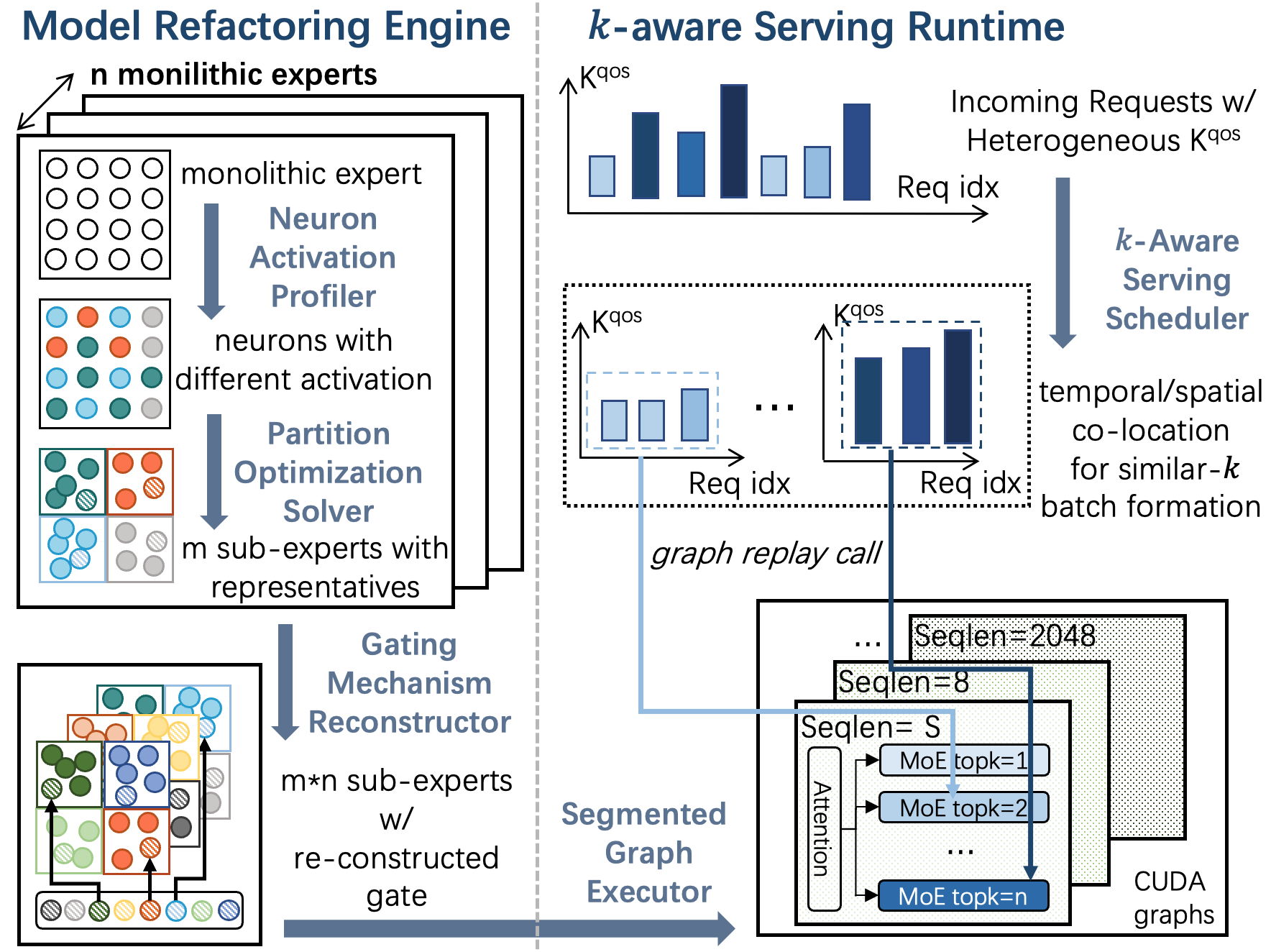}
    \caption{Overview of \modelname.}
    \label{fig:overview}
\end{figure}

Figure~\ref{fig:overview} shows the architecture of \modelname and how a request-level routing budget is carried through model refactoring and runtime execution. At deployment, the \emph{Model Refactoring Engine} partitions each pretrained expert into a set of finer-grained sub-experts and reconstructs the gate over the expanded expert set, exposing a denser spectrum of routing operating points. At serving time, the \emph{$k$-Aware Serving Scheduler} groups and dispatches requests with compatible budgets, preserving their computation savings without excessively fragmenting batches. Together, they turn top-$k$ into a practical runtime control knob for request-level compute elasticity.

Implementing this design efficiently in a modern serving stack introduces an additional engineering challenge: the expanded routing space substantially enlarges the CUDA-graph state space and its associated memory footprint. To address this issue, the \emph{Segmented Graph Executor} captures reusable attention segments across routing budgets while specializing MoE segments by $k$, limiting graph-memory growth while preserving CUDA-graph-based execution.

\subsection{Model Refactoring Engine}
\label{sec:model-side}

\subsubsection{Problem Formulation}

In mainstream MoE LLMs, experts are typically implemented with SwiGLU activation function. For input tokens $\mathbf{X}$, the intermediate activation of a SwiGLU expert is:
\begin{equation}
\mathbf{A}=\text{SiLU}(\mathbf{X}\cdot\mathbf{W}_{\text{gate}})\odot(\mathbf{X}\cdot\mathbf{W}_{\text{up}})
\end{equation}
and the FFN output is $\mathbf{X}'=\mathbf{A}\cdot\mathbf{W}_{\text{down}}$.

Each entry $\mathbf{A}_{p,d}$ depends exclusively on the $d$-th columns of $\mathbf{W}_{\text{gate}}$ and $\mathbf{W}_{\text{up}}$. We define a \emph{neuron} as the weights associated with one intermediate dimension: the $d$-th columns of $\mathbf{W}_{\text{gate}}$ and $\mathbf{W}_{\text{up}}$, alongside the $d$-th row of $\mathbf{W}_{\text{down}}$. Let $C$ be the intermediate size. The output $\mathbf{X}'$ is the sum of independent contributions from each neuron:
\begin{equation}
\mathbf{X}'_{p,q}=\sum_{d=0}^{C-1}\mathbf{A}_{p,d}\cdot\mathbf{W}^{\text{down}}_{d,q}.
\end{equation}

We partition the $C$ intermediate neurons into $N$ disjoint sets $\{S_1,\dots,S_N\}$, slicing the large weight matrices into $N$ smaller, independent sub-experts. (For simplicity, we assume that $C$ is divisible by $N$.) The original expert computation can then be rewritten as their accumulation:
\begin{equation}
  \mathbf{X}'_{p,q}
  =\sum_{n=1}^{N}\sum_{d\in S_n}
  \mathbf{A}_{p,d}\mathbf{W}^{\text{down}}_{d,q}.
\end{equation}
Here, $\{S_n\}_{n=1}^{N}$ is an equal-sized partition of the intermediate-neuron indices $\{0,\ldots,C-1\}$, with $|S_n|=C/N$.

The objective is to find a partition $\mathcal{P}$ such that, for most tokens, the activation mass is concentrated in only a few sub-experts, so dynamic top-$k$ execution can skip the rest with limited quality loss. The refactoring must also provide a router for the new sub-experts so that the resulting model remains compatible with existing deployment frameworks.

Formally, we run a calibration dataset through the pre-trained model and concatenate the activation batches of the calibration tokens routed to the current expert into $\mathbf{A}\in\mathbb{R}^{B\times C}$, where $B$ is the number of routed calibration tokens collected for that expert and $C$ is the intermediate size. Let $\mathcal{P}=\{S_1,\dots,S_N\}$ be a partition of the $C$ neurons into $N$ disjoint sub-experts. For token $b$, the activation mass of sub-expert $n$ is $L_{b,n}=\|\mathbf{A}[b,S_n]\|_1$. We use $L_{b,n}$ as a lightweight surrogate for the potential impact of skipping sub-expert $n$.
Let $k_{\mathrm{in}}$ denote the inner routing budget, i.e., the number of sub-experts retained within the current original expert.
The layer-wide runtime budget $k$ is the sum of such inner allocations across the selected original experts. 
Accordingly, $K_{\mathrm{drop}}=N-k_{\mathrm{in}}$ sub-experts are expected to be skipped.
Let $\phi_{K_{\mathrm{drop}}}(\cdot)$ denote the sum of the $K_{\mathrm{drop}}$ smallest elements.
The partition objective is
\begin{equation}
\mathcal{L}(\mathcal{P})=\sum_{b=0}^{B-1}\phi_{K_{\mathrm{drop}}}(\{L_{b,1},\dots,L_{b,N}\}),
\end{equation}
and the heuristic searches for a low-$\mathcal{L}$ partition $\widehat{\mathcal{P}}$. This concentrates activation mass in the remaining active sub-experts.

\subsubsection{Expert Decomposition}
To achieve the above goal, \modelname executes the refactoring in three steps. First, it uses \emph{Neuron Activation Profiler} to profile neuron activations on calibration data. Second, it solves the partitioning optimization to physically slice the monolithic experts into sub-experts with the \emph{Partition Optimization Solver}. Third, it reconstructs a lightweight gating mechanism to route tokens efficiently at runtime using \emph{Gating Mechanism Reconstructor}.

\begin{figure*}[t]
    \centering
    \includegraphics[width=0.9\linewidth]{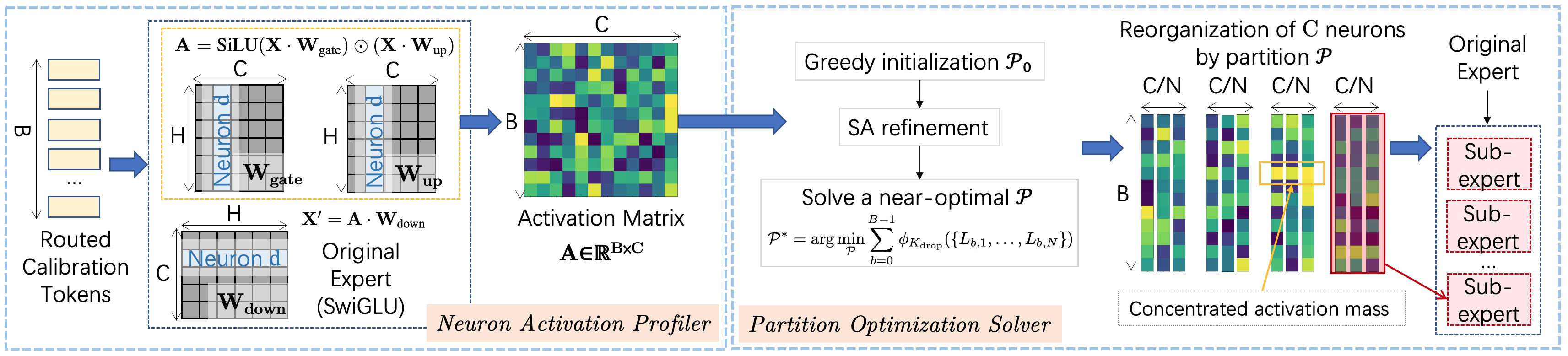}
    \caption{Overview of expert decomposition. Left: the \emph{Neuron Activation Profiler} leverages neuron-level independence in a SwiGLU FFN expert to collect the per-token activation value of each neuron. Right: the \emph{Partition Optimization Solver} uses the resulting activation matrix to group the original expert's neurons into fine-grained sub-experts according to their activation patterns.}
    \label{fig:how-carve}
\end{figure*}

\noindent\textbf{Neuron Activation Profiler.}
The \emph{Neuron Activation Profiler} (Fig.~\ref{fig:how-carve}, left) runs a calibration set through the pretrained model and records the intermediate activations of each expert. For each expert, the collected activations of its routed calibration tokens form an activation matrix $\mathbf{A}\in\mathbb{R}^{B\times C}$, where $B$ is the number of routed calibration tokens collected for that expert and $C$ is the intermediate size. The matrix $\mathbf{A}$ captures the per-token activation magnitude of each neuron which indicates the co-activation patterns among neurons. These activation matrices serve as the input to both partitioning and router reconstruction.

\noindent\textbf{Partition Optimization Solver.}
Given the activation matrix $\mathbf{A}$ of each expert, the \emph{Partition Optimization Solver} partitions each expert into fine-grained sub-experts (Fig.~\ref{fig:how-carve}, right). Since exact search is intractable, we use a two-stage heuristic. First, a greedy initializer ranks neurons by their calibration-wide $L_1$ norm and assigns each neuron to the eligible sub-expert with the smallest accumulated norm until every sub-expert contains $C/N$ neurons. Next, simulated annealing swaps neurons between groups, preserving equal sizes, and accepts better moves unconditionally and worse moves with a temperature-controlled probability, yielding the final partition maps $\{\widehat{\mathcal{P}}_e\}$.

\begin{figure*}[t]
    \centering
    \includegraphics[width=0.9\linewidth]{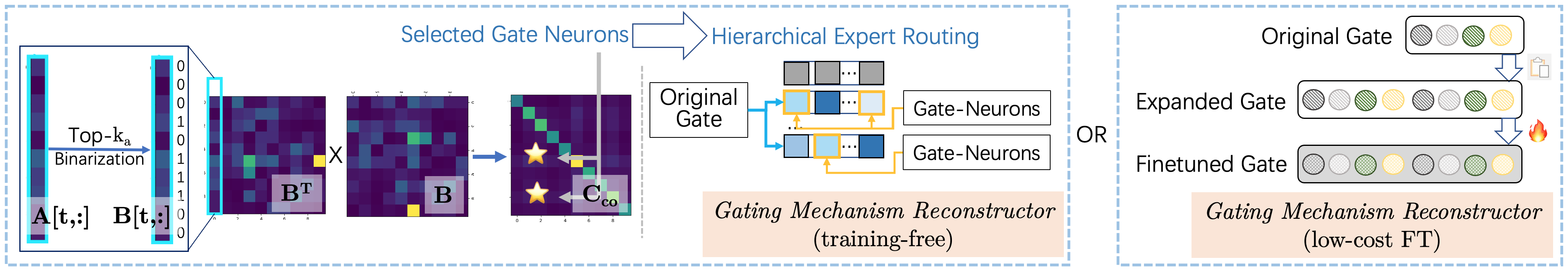}
    \caption{Overview of the two gating-mechanism reconstruction paths. Left: the training-free complex gate (CG) combines the original gate with representative gate neurons for hierarchical selection. Right: the expanded linear gate (LG) is initialized by repeating the original gate weights and then refined through gate-only fine-tuning.}
    \label{fig:how-gate}
\end{figure*}

\noindent\textbf{Gating Mechanism Reconstructor.}
After partitioning, we use the \emph{Gating Mechanism Reconstructor} to reconstruct a router for the new sub-experts. It provides two paths: a training-free proxy router for immediate deployment, and an expanded linear router with gate-only low-cost fine-tuning for maximum fidelity.

\noindent\emph{Training-Free Proxy Gating.}
For the training-free proxy router, we select gate neurons by co-activation (Fig.~\ref{fig:how-gate}, left). Given the calibration activation matrix $\mathbf{A}\in\mathbb{R}^{B\times C}$ from the profiler, we derive a binary matrix $\mathbf{B}\in\{0,1\}^{B\times C}$, where
\begin{equation*}
\mathbf{B}[t,c]=
\begin{cases}
1, & \text{if } c\in\operatorname{TopK}_{k_a}\bigl(|\mathbf{A}[t,:]|\bigr),\\
0, & \text{otherwise}.
\end{cases}
\end{equation*}
Here, $\operatorname{TopK}_{k_a}(\cdot)$ returns the indices of the $k_a$ largest entries. The co-activation matrix is then
\begin{equation*}
\mathbf{C}_{\text{co}}=\mathbf{B}^T\mathbf{B},
\end{equation*}
where $\mathbf{C}_{\text{co}}[i,j]$ counts how often neurons $i$ and $j$ are active together. For each sub-expert $S_n$, we score neuron $i\in S_n$ by its total co-activation with the other neurons in $S_n$ and keep the top-$r$ neurons as the gate-neuron set $G_n\subseteq S_n$, where $r$ denotes the number of gate neurons retained per sub-expert. At serving time, the pre-existing router first scores the original experts. Afterwards, for an input token $\mathbf{x}$ routed to a selected original expert, let
\begin{equation*}
a_c(\mathbf{x})=\operatorname{SiLU}(\mathbf{x}\mathbf{W}_{\mathrm{gate}}[:,c])
(\mathbf{x}\mathbf{W}_{\mathrm{up}}[:,c])
\end{equation*}
denote the activation of neuron $c$, where $\mathbf{W}_{\mathrm{gate}}[:,c]$ and $\mathbf{W}_{\mathrm{up}}[:,c]$ are the corresponding $c$-th columns of that expert's weight matrices. We compute $a_c(\mathbf{x})$ only for the selected gate neurons $G_n$ and use
\begin{equation*}
s_n(\mathbf{x})=\frac{1}{r}\sum_{c\in G_n}|a_c(\mathbf{x})|
\end{equation*}
as the proxy score of sub-expert $S_n$. Sub-experts derived from unselected original experts are not scored with this $s_n$. 

The original router selects its native number $m$ of original experts and orders them by router probability. Let $j$ be the index of a selected expert. For each evaluated layer-wide budget $k$, a configured vector $(\kappa_1,\ldots,\kappa_m)$ satisfies $0\le\kappa_j\le N$ and $\sum_j\kappa_j=k$, where $\kappa_j$ is the number of sub-experts allocated to original expert $j$. Among the $N$ sub-experts derived from original expert $j$, the proxy selects the top $\kappa_j$ by proxy score. Let $w_j$ be the normalized and scaled router weight of original expert $j$, every selected sub-expert of that expert inherits $w_j$ without a second softmax. The routed output is therefore $\sum_j w_j\sum_{n\in\operatorname{Top}_{\kappa_j}(s_j)}\mathrm{FFN}_{j,n}(\mathbf{x})$, where $s_j$ denotes the proxy scores of the sub-experts derived from original expert $j$. Selecting all $N$ sub-experts of an original expert reconstructs its original FFN contribution. The parameter $r$ controls the fidelity--overhead trade-off, and the added metadata scales only with the number of sub-experts rather than with the dominant FFN weights.

\noindent\emph{Low-Cost Fine-Tuning.}
For maximum fidelity, we support low-cost fine-tuning that updates only the router (Fig.~\ref{fig:how-gate}, right). To align the router with the expanded sub-expert space, we initialize the expanded router by repeating each row of the original gate linear weight, expanding its output dimension to the refactored expert count while preserving the original expert-level routing logits at initialization. We retain the model's routing normalization and apply the implementation's $N$-adjusted routed scaling. Unlike the hierarchical proxy path, this expanded linear router directly scores all refactored sub-experts and selects the layer-wide top-$k$. At the matched full budget, $k_{\mathrm{new}}=N k_{\mathrm{orig}}$, all $N$ sub-experts derived from each originally selected expert are activated. We optimize it on a small text corpus using the original language-modeling objective while freezing all non-router parameters. Following prior scalable MoE training~\cite{chen2023sparse}, we progressively increase the active sub-expert count toward the target $k$ during training instead of keeping it fixed. Because the reconstructed router remains a standard linear layer, it is directly compatible with existing MoE frameworks.

\subsection{$k$-Aware Serving Scheduler}
\label{sec:system-implementation}

\subsubsection{Problem Setup}
Having refactored the monolithic MoE model, we deploy it into a serving environment with externally specified routing budgets. The target $k_i^{\text{qos}}$ is supplied by a user, service tier, or upstream policy. Selecting this target is outside our scope.

We consider a system where requests arrive continuously and are executed in a batched manner. Each request is a tuple \(R_i = (a_i,\, n_i,\, k_i^{\text{qos}})\), where \(a_i\) is arrival time, \(n_i\) is the prompt token count, and \(k_i^{\text{qos}}\) is its requested routing budget. Offline serving operates on a finite bulk queue and optimizes throughput, whereas online serving receives streaming arrivals and prioritizes TTFT.

We use $k_{\mathrm{orig}}$ for an original whole-expert budget and $k_{\mathrm{sub}}=N k_{\mathrm{orig}}$ for its matched budget after each expert is split into $N$ sub-experts. Request target $k_i^{\mathrm{qos}}$ is measured in these sub-expert units. Intermediate values between adjacent $k_{\mathrm{sub}}$ points provide finer routing control.

A batch \(\mathcal{B}\) is constrained by a token budget \(\sum_{R_i \in \mathcal{B}} n_i \le M\). Because batched MoE kernels and graph replay rely on uniform routed-pair tensor shapes within one dispatch, once requests with different \(k_i^{\text{qos}}\) are mixed into the same dispatch, the efficient execution path typically uses the batch-level maximum routing budget, \(k(\mathcal{B}) = \max_{R_i \in \mathcal{B}} k_i^{\text{qos}}\), so that every request receives at least its requested quality budget. This \emph{over-serves} requests with \(k_i^{\text{qos}} < k(\mathcal{B})\). Although the runtime could split an already mixed batch into multiple smaller dispatches, doing so generally sacrifices batching efficiency. Therefore, \modelname instead avoids forming highly heterogeneous dispatches in the first place by reordering and distributing requests with similar routing budgets together.

The scheduler works in a non-preemptive decode policy: once a request enters decode, we do not evict or migrate it because decode interruption would inflate time per output token (TPOT) p99. Therefore, scheduling decisions apply only to requests that have not started running. To minimize over-serving while respecting \(k_i^{\text{qos}}\), the scheduler employs two complementary strategies: \emph{temporal separation} for throughput-oriented offline serving, and \emph{spatial separation} for latency-sensitive online serving.

\begin{figure*}[t]
    \centering
    \begin{minipage}[t]{0.49\textwidth}
        \centering
        \includegraphics[width=0.85\linewidth]{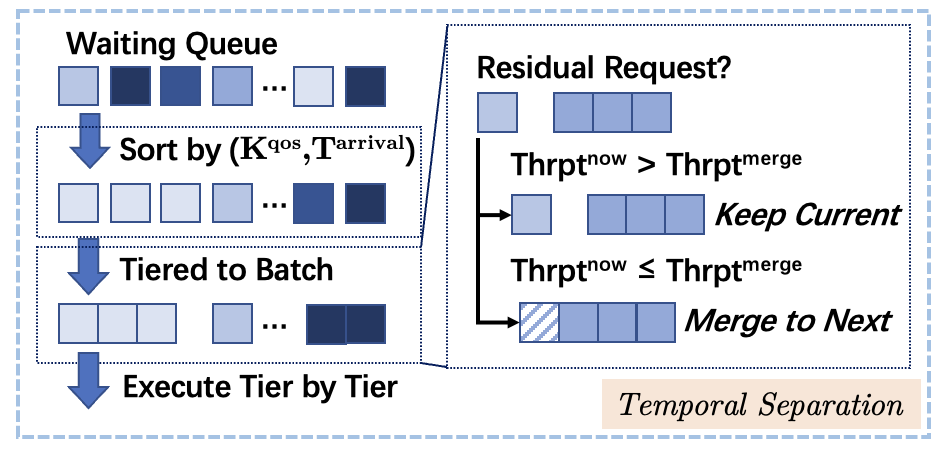}
        \caption{Temporal separation. Offline serving uses quality-tiered batching.}
        \label{fig:offline}
    \end{minipage}
    \hfill
    \begin{minipage}[t]{0.49\textwidth}
        \centering
        \includegraphics[width=0.85\linewidth]{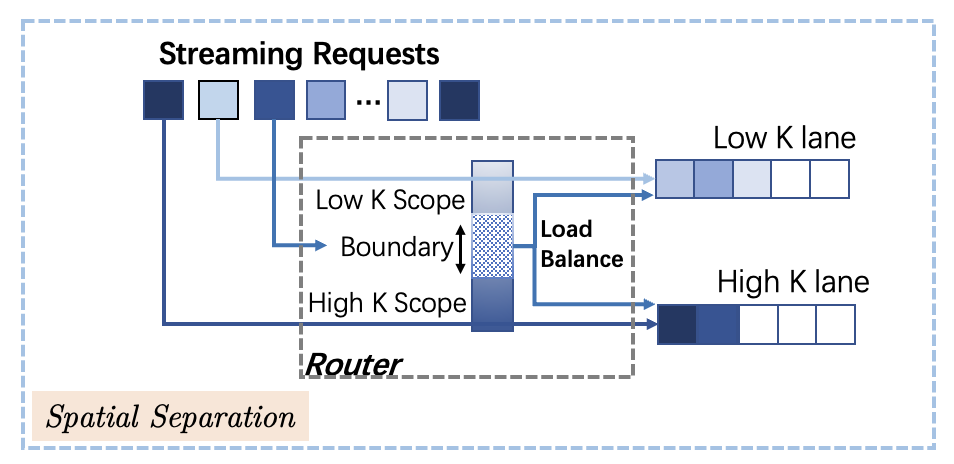}
        \caption{Spatial separation. Online serving uses lane-based routing with an adaptive boundary.}
        \label{fig:online}
    \end{minipage}
\end{figure*}

\subsubsection{Temporal Separation via Quality-Tiered Batching}
\label{sec:offline-sche}

For throughput-oriented workloads, the scheduler minimizes over-serving by batching requests with similar \(k^{\text{qos}}\). (Fig.~\ref{fig:offline}) The waiting queue is sorted by \((k_i^{\text{qos}}, a_i)\), prioritizing lower-quality requests while maintaining FIFO order within each tier. The scheduler scans tiers in ascending order to construct batches.
When a tier does not entirely fill the budget \(M\), a residual batch remains. The scheduler faces a trade-off: dispatch the residual batch immediately at its current quality \(k\) or merge it with the next tier \(k_{\mathrm{next}}>k\). The former avoids over-serving but under-utilizes hardware, whereas the latter improves batch fullness at the expense of over-serving residual requests.

To resolve this trade-off, the scheduler performs a local throughput comparison at each adjacent boundary. Let $n_{\mathrm{cur}}$ be the prompt tokens already accumulated in the residual batch at budget $k$, let $n_{\mathrm{next}}$ be the tokens contributed by the next tier under the remaining budget, and let $T(n,k)$ be the profiled execution time for processing $n$ prompt tokens at budget $k$. The scheduler then checks
\begin{equation}
\frac{n_{\mathrm{cur}}+n_{\mathrm{next}}}{T(n_{\mathrm{cur}}+n_{\mathrm{next}}, k_{\mathrm{next}})}
\ge
\frac{n_{\mathrm{cur}}}{T(n_{\mathrm{cur}}, k)}.
\end{equation}
If the inequality holds, the residual batch is promoted to budget $k_{\mathrm{next}}$ and merged with the next tier. Otherwise, it is dispatched immediately at budget $k$. Thus, over-serving is introduced only when it improves token throughput through better batch fullness.

This is a local, hardware-related heuristic rather than a global scheduling optimum. The table \(T(n,k)\) is profiled at representative prefill-token and supported-$k$ buckets on the deployment GPU, with interpolation for unprofiled points. A misprediction only affects the residual part of one under-filled batch and only across one adjacent quality tier. Subsequent batches are rebuilt from the live queue, so such errors do not accumulate across scheduling decisions.

\subsubsection{Spatial Separation via Lane-Based Routing}
\label{sec:online-sche}

When requests must be served online, temporal separation is tightly constrained by TTFT. We rely on spatial separation in a data-parallel deployment with multiple DP ranks (Fig.~\ref{fig:online}). We build on vLLM's original data-parallel dispatcher, which assigns each request to the DP rank with minimum queue-pressure score $\mathrm{Score}_j = 4W_j + R_j$, where $W_j$ and $R_j$ are the waiting and running counts of DP rank $j$. This policy balances queue pressure well, but it is agnostic to $k_i^{\text{qos}}$ and may mix very different requests on the same rank, increasing over-serving.

Rigidly partitioning ranks by quality is unsuitable because bursty arrivals can overload one subset while leaving another under-utilized, directly hurting TTFT. The key trade-off is therefore between reducing over-serving and avoiding queue buildup. To navigate this trade-off, the scheduler partitions DP ranks into low-$k$ and high-$k$ lanes and maintains a soft boundary window $[b_{\mathrm{low}},\,b_{\mathrm{high}}]$ spanning $w$ consecutive budget levels around $b$, clipped to the supported range $[k_{\min},k_{\max}]$. For request $R_i$, if $k_i^{\text{qos}} < b_{\mathrm{low}}$, it prefers the low-$k$ lane; if $k_i^{\text{qos}} > b_{\mathrm{high}}$, it prefers the high-$k$ lane; otherwise it is eligible for either lane and will be dispatched by the load balance manner. Requests falling within the window act as flexible buffers to absorb short-term load imbalance.

Within a preferred lane \(\ell\), we reuse vLLM's original queue-aware rank-selection rule, which selects the minimum-score rank:
\begin{equation}
j^* = \arg\min_{j \in \mathcal{C}_{\ell}(R_i)} \left(4W_j + R_j\right),
\end{equation}
where $\mathcal{C}_{\ell}(R_i)$ is the eligible rank set. Hence, the design preserves vLLM's queue-aware rank selection while injecting $k$-awareness at lane selection. If the preferred lane has no idle rank but another lane does, the scheduler may bypass the preference to avoid obvious capacity waste.

To handle sustained workload skew, the boundary adapts online based on lane-level pressure \(P_{\ell} = 4W_{\ell} + R_{\ell}\), where $W_{\ell}$ and $R_{\ell}$ are the aggregated waiting and running counts of lane $\ell$. If the low-$k$ lane is substantially lighter, the boundary moves upward; if the high-$k$ lane is lighter, it moves downward. Hysteresis and cooldown prevent oscillation, and the hysteresis grows as the boundary drifts farther from its initial value. This soft lane-based design reduces unnecessary expert computation without sacrificing TTFT under bursty online arrivals.

\subsection{Implementation in \texttt{vLLM}}
\label{sec:implementation}

\subsubsection{From Expert Refactoring to Runtime Pressure}

The key issue is that the model-side refactoring and the runtime are tightly coupled. On the model side, expert refactoring decomposes each original expert into finer-grained sub-experts and turns a small set of coarse routing choices into a denser set of elastic $k$ operating points. This extra routing flexibility is exactly what enables dynamic $k$ serving. On the system side, however, the runtime must now support many more routing states efficiently.

Modern high-performance serving systems such as \texttt{vLLM} rely on CUDA Graphs to amortize CPU-side kernel-launch overhead and optimize throughput and latency. Disabling graphs would make elastic $k$ easier to support functionally, but would also sacrifice a core efficiency mechanism of the serving system. Therefore, the runtime must preserve graph-based execution even after expert refactoring enlarges the routing space.

In standard executors, cached graphs are keyed primarily by sequence shapes (\(S\) buckets). Because elastic \(k\) alters the volume of routed token-expert pairs, the routing budget (\(K\) operating points) becomes a new graph specialization dimension. Consequently, a monolithic graph design would require up to \(S \times K\) cached graphs. Since captured CUDA graphs consume non-trivial GPU memory, this multiplicative growth would quickly crowd out KV-cache capacity, negating the practical benefits of expert refactoring.

\subsubsection{The Segmented Graph Executor}
\label{sec:executor}

\begin{figure}[t]
    \centering
    \includegraphics[width=0.8\linewidth]{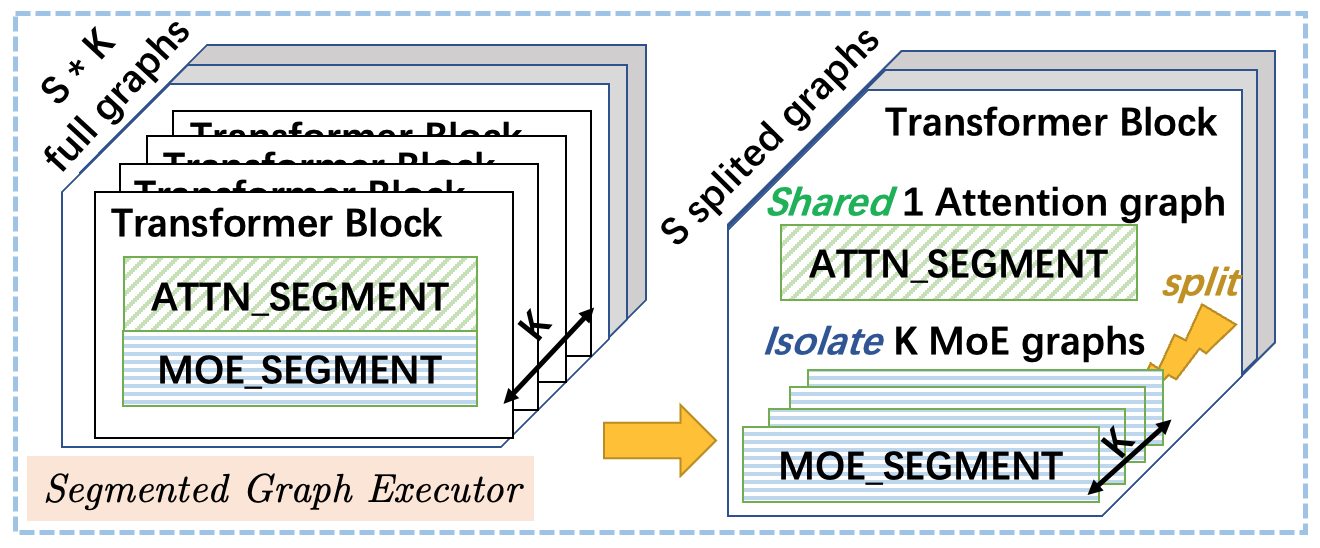}
    \caption{We split MoE and Attention layer in the Segmented Graph Executor. 
    }
    \label{fig:graph}
\end{figure}

To retain CUDA-graph replay without paying the full graph-cache cost, we exploit a structural asymmetry inside each Transformer block. The attention path depends on sequence-related shapes, but it is insensitive to the routing budget. In contrast, the MoE path is directly routing-sensitive, because changing $k$ changes the amount of expert computation that must be executed. Our executor (Fig. \ref{fig:graph}) therefore specializes only the MoE portion by routing budget while allowing the attention portion to be shared across different $k$ values. This is different from vLLM's existing piecewise graph mode, which targets graph-incompatible operations by keeping them eager while capturing the remaining graph-compatible operations. In our setting, both attention and MoE are graph-compatible, but only MoE needs $k$ specialization. The existing piecewise mode is therefore not sufficient to share attention graphs across routing budgets while caching separate MoE graphs for each $k$.

To realize this idea, we modify the graph-capture path to split each Transformer block at the Attention-MoE boundary instead of capturing the whole block as one graph. During capture, every MoE invocation is treated as a mandatory cut point, producing two segment classes: \texttt{ATTN\_SEGMENT} for the routing-insensitive attention path and \texttt{MOE\_SEGMENT} for the routing-sensitive expert path. Each segment is then captured independently together with its shape metadata and segment type.

At replay time, cache lookup uses the token count for both segment types but handles the batch's execution budget $k(\mathcal{B})$ differently.
For \texttt{MOE\_SEGMENT}s, the budget is preserved in the lookup key, while for \texttt{ATTN\_SEGMENT}s, it is masked before lookup. Batches with the same token count but different routing budgets can therefore reuse attention graphs, while execution alternates between shared attention and $k$-specialized MoE segments in program order.

Finally, we integrate elastic $k$ throughout vLLM's serving pipeline. Each request carries its budget $k$ in its metadata, which is propagated to the input batch. The maximum $k$ in the batch then configures each \texttt{fused\_moe\_kernel} call. To make these choices replayable, graph capture and warm-up instantiate the MoE segments for all configured budgets before serving. In this way, \modelname preserves the efficiency benefits of CUDA-graph-based serving while exposing a much richer set of routing operating points to the scheduler.

\section{Experimental Setup}

\subsection{Test Models}

We select three open-source MoE models (Table~\ref{tab:baseline_comparison1}) to demonstrate the general applicability of \modelname. We expand the native range of $k_{\mathrm{orig}}$ by a factor of four to obtain $k_{\mathrm{sub}}$, yielding four times as many operating points. The activation matrix $\mathbf{A}$ required for partitioning optimization is collected by running calibration on the \texttt{WikiText-2-raw-v1} dataset train split~\cite{merity2016pointer}. The expert partitioning process leverages a simulated annealing algorithm configured with an initial temperature $T_0=100.0$, a cooling rate $\alpha=0.995$, and for $I=100,000$ iterations.

\begin{table}[h]
\centering
\caption{Architectural parameters of the baseline models.}
\label{tab:baseline_comparison1}
\begin{tabular}{lccc}
\toprule
\textbf{Attribute} & \texttt{OLMoE-P} & \texttt{DeepSeek-P} & \texttt{Qwen-P} \\
\midrule
Total Parameters & 7B & 16B & 30B \\
Routing Experts & 64$\rightarrow$256 & 64$\rightarrow$256 & 128$\rightarrow$512  \\
Activated Experts & 8$\rightarrow$32 & 6$\rightarrow$24 & 8$\rightarrow$32   \\
Intermediate Size & 1024$\rightarrow$256 & 1408$\rightarrow$352 & 6144$\rightarrow$1536\\
\bottomrule
\end{tabular}

\vspace{2pt}
  \parbox{\columnwidth}{%
    \emph{Note:} \footnotesize OLMoE-P is generated from allenai/OLMoE-1B-7B-0924, DeepSeek-P is generated from deepseek-ai/deepseek-v2-lite, and Qwen-P is generated from Qwen/Qwen3-30B-A3B.
    }
\end{table}

For the training-free gate reconstruction, we identify the top $r=4$ neurons from each sub-expert as gate neurons. These neurons are selected by our algorithm from the co-activation matrix derived from the top $k_a=\frac{3}{4} C$ activated neurons for each token in the calibration set, where $C$ denotes the total number of neurons in the corresponding FFN layer. We keep the $r$ value fixed to isolate the impact of expert refactoring and runtime scheduling. Additionally, we test gate-only fine-tuning with a learning rate of $1 \times 10^{-5}$, a linear annealing of $k$ from 8 to 24 (for \texttt{DeepSeek}) or 32 (for \texttt{OLMoE} and \texttt{Qwen}), and an accumulated batch size of 32 (for \texttt{DeepSeek} and \texttt{Qwen}) or 64 (for \texttt{OLMoE}). We use 200K sampled sequences from the \texttt{SlimPajama} dataset~\cite{cerebras2023slimpajama} as the training set.

This refactoring workflow is a one-time deployment-time preprocessing step rather than a serving-path cost. The refactoring requires calibration, partitioning, and gate reconstruction. We therefore treat the refactoring cost as amortized over the deployment lifetime and focus the runtime evaluation on serving-time throughput and latency.

\subsection{System Setup and Metrics}

We implement \modelname on \texttt{vLLM} v0.11.2 with CUDA 12.6, Python 3.11, and PyTorch 2.9.0. We use \texttt{vLLM}'s standard serving and benchmarking interfaces for request generation and metric collection.

\subsubsection{Workloads for Execution Accuracy}

We assess model quality using the Eleuther AI Language Model Evaluation Harness (\texttt{lm-eval})~\cite{eval-harness} with \texttt{vLLM} as inference backend. We report perplexity on the \texttt{WikiText} dataset~\cite{merity2016pointer} and accuracy on \texttt{WinoGrande} (3-shot), \texttt{ARC-Challenge} (5-shot), \texttt{SciQ} (0-shot), and \texttt{BoolQ} (0-shot)~\cite{ai2winogrande,clark2018think,welbl2017crowdsourcing,clark2019boolq}.

\subsubsection{Workloads for Execution Efficiency}
\label{sec:expri:set:eff}

For serving throughput and latency evaluation, we use the Azure serving trace released by DynamoLLM~\cite{stojkovic2025dynamollm}. 
We filter the trace to retain requests whose total token length is between 1024 and 2048 tokens to match the capacity regime of our testbed. 
We use the filtered trace to preserve realistic request-size characteristics, while controlling request arrivals as a Poisson process with a configurable rate. 
This setup enables us to evaluate \modelname under realistic request-shape distributions while varying the offered load in a controlled manner.

For elastic serving experiments, each request is additionally assigned a routing-budget target $k_i^{\text{qos}}$.
We generate $k_i^{\text{qos}}$ from a truncated normal distribution:
\begin{equation}
k_i^{\text{qos}} = \mathrm{clip}\!\left(\mathrm{round}(x_i),\, 1,\, k_{\max}\right),
\quad x_i \sim \mathcal{N}(\mu_k, \sigma_k^2),
\end{equation}
where $\mu_k$ and $\sigma_k$ are generator parameters, and $k_{\max}$ is the model's maximum sub-expert budget. These synthetic targets serve as a systems stress workload, emulating the heterogeneous routing budgets that could be derived from request-level quality requirements or SLOs. Unless stated otherwise, all serving systems use the same refactored LG w /FT checkpoint, request sequence, and seed.

\section{Evaluation Results}

\subsection{Offline Task Throughput Performance}

\begin{figure*}[t]
    \centering
    \includegraphics[width=1\linewidth]{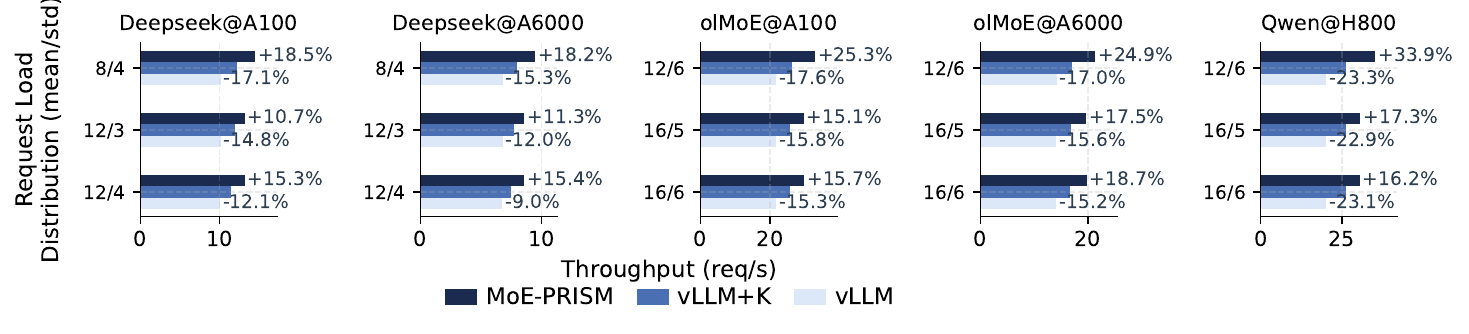}
    \caption{End-to-end offline throughput (req/s) under routing target generator parameters $\mu_k/\sigma_k$. \texttt{vLLM} uses fixed $k_{\max}$; \texttt{vLLM+K} adds dynamic-$k$ execution with FIFO; \modelname additionally uses the $k$-Aware Serving Scheduler. Percentage annotations use \texttt{vLLM+K} as denominator.}
    \label{fig:offline_res}
\end{figure*}

We evaluate the end-to-end benefit of \modelname in throughput-oriented serving, where per-request latency is not part of the optimization objective. In this setting, the full system should translate request-level $k$ heterogeneity into higher throughput by combining elastic $k$ execution with the offline scheduler described in Section~\ref{sec:offline-sche}.

All three systems use the same request sequence and random seed. \texttt{vLLM} uses the original, unrefactored model, ignores request targets, and runs fixed $k_{\max}$ with its FIFO scheduler. \texttt{vLLM+K} and \modelname use the same refactored LG w/ FT checkpoint and segmented graph path; \texttt{vLLM+K} adds our dynamic-$k$ execution extension while retaining FIFO, and \modelname further adds the offline scheduling policy. We report end-to-end request throughput (req/s) under workloads with different $k_i^{\text{qos}}$ distributions, controlled by the routing target generator parameters $\mu_k$ and $\sigma_k$.

Fig.~\ref{fig:offline_res} summarizes the results across DeepSeek-V2-Lite on A100/RTX A6000, OLMoE-1B-7B on A100/RTX A6000, and Qwen3-30B-A3B on H800. Moving from \texttt{vLLM} to \texttt{vLLM+K} already improves throughput by exposing request-level routing heterogeneity, and \modelname further improves throughput by scheduling requests with similar routing budgets together. These results confirm that the complete runtime stack can translate elastic routing control into end-to-end serving gains.

\subsection{Online Task Latency Performance}

\begin{figure*}[t]
  \centering
  \includegraphics[width=0.98\linewidth]{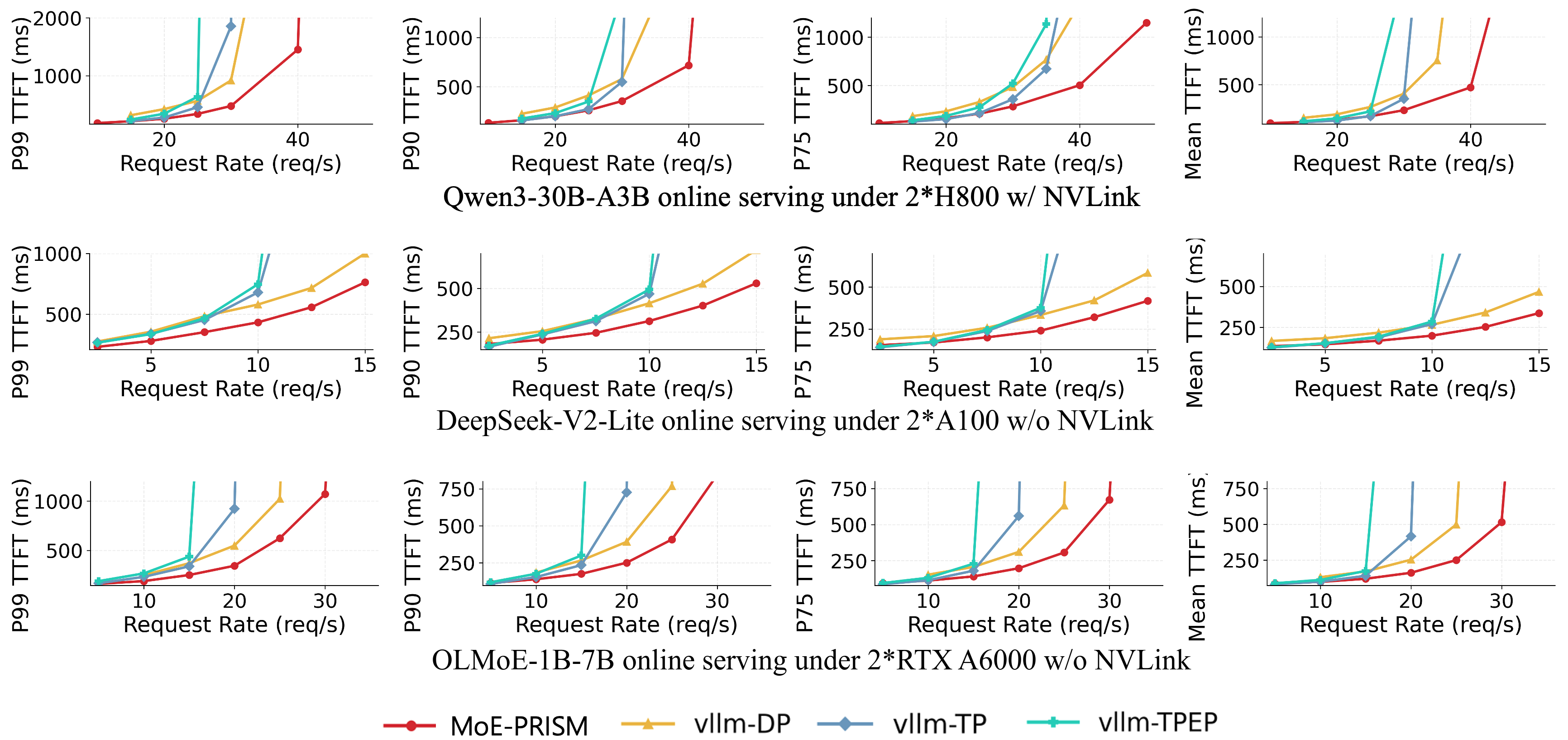}
  \caption{Full-system online TTFT under increasing offered load. Rows use Qwen on 2$\times$H800 with $(\mu_k,\sigma_k)=(16,5)$, DeepSeek on 2$\times$A100 with $(12,3)$, and OLMoE on 2$\times$RTX A6000 with $(16,5)$. \modelname is compared against the three \texttt{vLLM}-based baselines.}
  \label{fig:double}
\end{figure*}

We next evaluate the end-to-end benefit of \modelname under latency-sensitive serving  described in Section~\ref{sec:online-sche}. In this setting, the system must reduce unnecessary MoE computation while still keeping queueing delay under control across DP ranks.

We compare \modelname's online runtime against three \texttt{vLLM}-based baselines, reported in the figure legend as \texttt{vllm-DP} (data parallel), \texttt{vllm-TP} (tensor parallel), and \texttt{vllm-TPEP} (tensor parallel for attention and expert parallel for MoE). 
Every request is assigned a routing-budget target $k_i^{\text{qos}}$ sampled from the truncated normal distribution defined in \S~\ref{sec:expri:set:eff}. 
These baselines use the original, unrefactored models and their native fixed routing budgets. In contrast, \modelname uses the refactored LG w/FT models with DP, consumes the per-request $k$ targets, and enables both dynamic-$k$ execution and the $k$-Aware Serving Scheduler. The scheduler-only comparison against dynamic-$k$ \texttt{vLLM+K} is reported separately in the ablation section (\S~\ref{sec:abla:scheduler}). Fig.~\ref{fig:double} reports mean, p75, p90, and p99 TTFT under increasing request rates for three test environments.

Across these settings, \modelname generally achieves the lowest measured TTFT and delays the onset of sharp queueing growth to higher offered loads. The differences become more visible near saturation loads, where the baselines' tail latency often rises earlier. Overall, these results indicate that the full online design can exploit request-level routing heterogeneity to sustain lower latency as the system approaches saturation.

\subsection{Execution Accuracy}

\begin{figure}[t]
\centering
\includegraphics[width=1\linewidth]{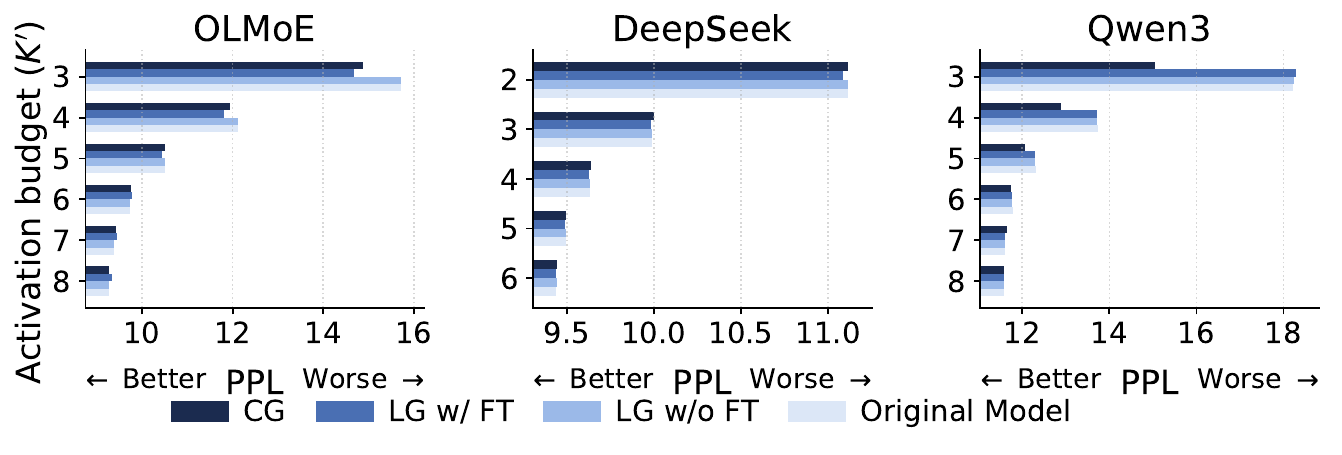}
\caption{Absolute WikiText perplexity at matched budgets. The original model uses $k_{\mathrm{orig}}$, whereas each refactored model uses $k_{\mathrm{sub}} = 4k_{\mathrm{orig}}$. CG is the training-free complex gate. LG w/o FT and LG w/ FT are the expanded linear gate before and after gate-only fine-tuning.}
\label{fig:model_ppl}
\end{figure}

We evaluate our proposed method against the original models at matched activated  parameters to provide a fair and informative assessment of the model-refactoring mechanism described in Section~\ref{sec:model-side}. 
Fig.~\ref{fig:model_ppl} reports the absolute perplexity on \texttt{WikiText}, where lower values indicate better model quality. Overall, our model refactoring preserves the accuracy of the original model. Across the evaluated activation budgets, the training-free complex gate and the linear gate variants before and after gate-only fine-tuning remain close to their corresponding original baselines, and several configurations achieve modestly lower perplexity. These improvements are particularly evident at lower activation budgets for OLMoE and Qwen3, suggesting that the refactored expert structure can sometimes support more effective expert selection under constrained computation.
Table~\ref{tab:benchmark_results} presents the downstream task results for the linear router after gate-only fine-tuning. Across the evaluated settings, the refactored variants are generally competitive with their original counterparts and occasionally achieve slightly better performance. 
Taken together, the perplexity and downstream-task results suggest that our refactoring mechanism largely preserves the capabilities of the original models rather than introducing a substantial quality penalty. At the same time, it exposes a finer-grained set of $k$ control points, enabling more flexible adjustment of the activation budget and providing additional operating points for navigating the trade-off between model quality and inference computation cost.

\begin{table}[t]
\centering
\caption{Downstream accuracy at matched budgets.}
\label{tab:benchmark_results}
\begin{tabular}{lcccc}
\toprule
\textbf{Config} & \textbf{WinoGrande} & \textbf{ARC-C} & \textbf{SciQ}   & \textbf{BoolQ} \\
\midrule 
OLMoE $k_{\mathrm{orig}}=3$ & 63.06 & 46.08 & 87.60 & 64.46 \\
OLMoE-P $k_{\mathrm{sub}}=12$ & \cellcolor{gainbg}+1.27 & \cellcolor{gainbg}+0.25 & \cellcolor{gainbg}+2.10 & \cellcolor{gainbg}+3.19 \\
\midrule 
DeepSeek $k_{\mathrm{orig}}=3$ & 72.85 & 52.73 & 93.10 & 75.93 \\
DeepSeek-P $k_{\mathrm{sub}}=12$ & \cellcolor{gainbg}+0.79 & +0.00 & \cellcolor{dropbg}-0.10 & \cellcolor{gainbg}+0.13 \\
\midrule 
Qwen $k_{\mathrm{orig}}=3$ & 58.17 & 51.62 & 84.70 & 70.70 \\
Qwen-P $k_{\mathrm{sub}}=12$ & \cellcolor{dropbg}-0.47 & \cellcolor{gainbg}+2.48 & \cellcolor{gainbg}+0.40 & \cellcolor{gainbg}+0.06 \\
\bottomrule
\end{tabular}

\vspace{2pt}
  \parbox{\columnwidth}{%
    \footnotesize
    \emph{Note:}  Original rows are absolute percentages. OLMoE-P/DeepSeek-P/Qwen-P rows use refactored models with LG w/ FT and report percentage-point changes relative to the corresponding original rows.
    }

\end{table}

\subsection{Ablation Study}

\subsubsection{Impact of Expert Refactoring}
We first study the effect of expert refactoring on dynamic top-$k$ serving. The main benefit of refactoring is that it reduces the granularity mismatch between the request-level $k_i^{\text{qos}}$ target and the actual routing unit exposed by the model. In the refactored setting, each original expert is split into multiple fine-grained sub-experts, allowing the runtime to realize a request's target routing budget more precisely. In the original coarse-grained setting, however, routing can only be performed at the granularity of full experts, which forces the system to round up fine-grained routing targets and may introduce unnecessary computation.

Formally, $k_i^{\text{qos}}$ is measured in routed sub-expert units excluding shared experts. The original layout must execute
$ k_i^{\text{exec}} = N \left\lceil \frac{k_i^{\text{qos}}}{N} \right\rceil $
in the same units. With $N=4$, targets that are not multiples of $4$ are rounded up, producing \emph{granularity-induced over-serving}: even with an ideal scheduler, the coarse-grained system may still do more computation than required to meet the request target.

Table~\ref{tab:ablation_refactor} compares the original coarse-grained layout and the refactored fine-grained layout under identical targets and the same offline scheduler. The results show that expert refactoring consistently reduces routing-budget overshoot and improves execution efficiency, confirming that fine-grained expert decomposition is an important enabler for request-level elastic $k$ serving.

\begin{table}[h]
\centering
    \caption{Impact of expert refactoring on 1,024 OLMoE offline requests on one RTX A6000.}

\label{tab:ablation_refactor}
\begin{tabular}{cccc}
\toprule
\textbf{$\mu_k$} & \textbf{$\sigma_k$} & \textbf{Orig. Model} & \textbf{Refac. Model} \\
\midrule
16  & 8 & 25.38 / 20.44 & 26.06 / 17.17 \\
16  & 6 & 24.54 / 18.36 & 25.97 / 16.76 \\
12  & 8 & 27.10 / 14.68 & 28.68 / 12.95 \\
\bottomrule
\end{tabular}

\vspace{2pt}
  \parbox{\columnwidth}{%
    \footnotesize
    \emph{Note:} Each entry reports throughput (req/s) / the token-weighted mean executed batch budget expressed in routed sub-expert units. $\mu_k$ and $\sigma_k$ parameterize the generator of request targets $k_i^{\mathrm{qos}}$.
  }
\end{table}

\subsubsection{Impact of the Segmented Graph Executor}

We next study the cost-benefit trade-off of the Segmented Graph Executor. Its purpose is to reduce the graph-cache overhead introduced by elastic $k$ serving, but it also changes the execution path by splitting the original monolithic graph into multiple segments and introducing segment-level graph dispatch. Therefore, the key question is whether the memory savings justify the additional runtime indirection and whether the extra Attention-MoE boundaries noticeably hurt throughput.

We compare four configurations: (1) the original model with the default $k$-agnostic graph executor, (2) the original model with naive $k$-aware graph dispatch, (3) the refactored model with naive $k$-aware graph dispatch, and (4) the refactored model with our Segmented Graph Executor. Table~\ref{tab:ablation_cudagraph} reports the graph-related memory usage with the default vLLM capture-size buckets and the throughput under the maximum supported routing budget.

\begin{table}[h]
\centering
\caption{Impact of the Segmented Graph Executor for 1,024 OLMoE requests on one RTX A6000.}
\label{tab:ablation_cudagraph}
\begin{tabular}{lcc}
\toprule
\textbf{Configuration} & \textbf{Graph Mem. (GB)} & \textbf{Thrpt (req/s)} \\
\midrule
Orig., $k$-agnostic & 0.27 & 20.81 \\
Orig., $k$-aware & 1.56 & 20.82 \\
Refac., $k$-aware & 6.19 & 19.97 \\
\modelname & 0.96 & 19.93 \\
\bottomrule
\end{tabular}

\vspace{2pt}
  \parbox{\columnwidth}{%
    \footnotesize
    \emph{Note:} No scheduler is involved for all configurations evaluated at fixed $k$.
  }
\end{table}

The results show two trends. First, making the executor $k$-aware increases memory usage even for the original model, since the runtime must cache separate graph instances for different routing budgets. This effect becomes substantially worse after expert refactoring, where the finer-grained routing space further expands the number of graph states. Second, after enabling the Segmented Graph Executor, the graph-memory footprint is reduced markedly, while the throughput at maximum $k$ remains nearly unchanged compared with the naive $k$-aware execution: throughput changes only from 19.97 to 19.93 requests/s, a 0.2\% drop. This indicates that the extra split between attention and MoE introduces only negligible serving-time overhead, and that the additional segment-level dispatch does not materially reduce serving efficiency. Overall, the Segmented Graph Executor effectively trades a negligible execution overhead for a substantial reduction in graph-related memory cost.

\subsubsection{Impact of Scheduling under Elastic $k$ Workloads}
\label{sec:abla:scheduler}

We next isolate the contribution of our scheduler for elastic $k$ workloads. The elastic $k$ runtime is fixed throughout this ablation; what changes is only how requests with different routing budgets are grouped and placed. In both the offline and online settings, we compare \texttt{vLLM+K}, namely the original \texttt{vLLM} scheduler/load balancer with elastic $k$ execution, against our scheduler. This ablation therefore asks whether reduced per-request computation alone is sufficient, or whether the system also needs routing-budget-aware scheduling to realize the benefit.

\begin{figure}[t]
    \centering
    \includegraphics[width=1\linewidth]{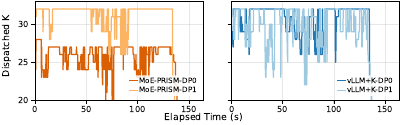}
    \caption{OLMoE routing-budget trace at 30 req/s on 2$\times$RTX A6000, with target generator $(\mu_k,\sigma_k)=(16,8)$. Left: \modelname separates lower- and higher-$k$ lanes. Right: \texttt{vLLM+K} mixes both regimes, increasing cross-$k$ interference and over-serving.}
    \label{fig:olmoe_online}
\end{figure}

Fig.~\ref{fig:olmoe_online} shows a representative online trace for OLMoE at 30 req/s. With our online scheduler, the two DP lanes separate into a lower-$k$ lane and a higher-$k$ lane, while \texttt{vLLM+K} causes both lanes to hover in a similar high-$k$ range, indicating persistent cross-$k$ interference. This cleaner lane specialization lets the system send requests with similar routing budgets to the same lane, so low-$k$ requests are less likely to be mixed with high-$k$ requests and over-served by the batch's larger routing budget. As summarized in Table~\ref{tab:ablation_kaware_online}, the online scheduler further reduces TTFT and TTFT p99 beyond what elastic $k$ alone can provide. These results show that elastic $k$ does not fully remove over-serving, and explicit scheduling for heterogeneous routing budgets is needed to turn routing-budget heterogeneity into end-to-end performance gains.

\begin{table}[htbp]
\centering
\caption{Scheduler-only online gain for OLMoE at 30 req/s with target generator $(\mu_k,\sigma_k)=(16,8)$.}
\label{tab:ablation_kaware_online}
\begin{tabular}{lccc}
\toprule
\textbf{\quad Method \quad} & \textbf{Mean TTFT} & \textbf{TTFT p99} & \textbf{Mean TPOT} \\
\midrule
\texttt{vLLM+K} & 229.15 & 572.44 & 82.42 \\
+ Scheduler & 222.24 & 526.71 & 80.16 \\
\bottomrule
\end{tabular}
\end{table}

\begin{figure}[t]
    \centering
    \includegraphics[width=1\linewidth]{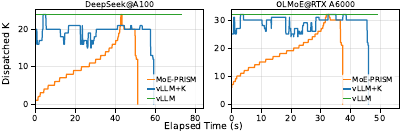}
    \caption{Offline dispatched-$k$ traces for DeepSeek on one A100, $(\mu_k,\sigma_k)=(12,5)$ (left), and OLMoE on one RTX A6000, $(16,6)$ (right). \modelname prevents the persistent high-\(k\) envelope caused by mixed batches.}
    \label{fig:dispatch-k-two-panels}
\end{figure}

Fig.~\ref{fig:dispatch-k-two-panels} shows two representative offline traces. Under \texttt{vLLM+K}, the dispatched routing budget stays close to the upper envelope of co-arriving requests, because low-$k$ and high-$k$ requests are still mixed in the same batch. In contrast, our offline scheduler produces a low-to-high progression and spends substantially more time at lower dispatched $k$, indicating that it forms batches with more homogeneous routing budgets and avoids residual over-serving. Table~\ref{tab:ablation_kaware_offline} confirms that this trace-level effect translates into sizable throughput gains on both workloads.

\begin{table}[htbp]
\centering
\caption{Scheduler-only offline gain (req/s) from adding the $k$-Aware Serving Scheduler on top of elastic top-$k$. }
\label{tab:ablation_kaware_offline}
\begin{tabular}{lccc}
\toprule
\textbf{Workload} & \textbf{vLLM+K} & \textbf{+ Scheduler} & \textbf{\quad Gain \quad} \\
\midrule
OLMoE-P 16/6& 21.21 & 26.06 & +22.9\% \\
DeepSeek-P 12/5& 15.30 & 17.36 & +13.5\% \\
\bottomrule
\end{tabular}

\vspace{2pt}
  \parbox{\columnwidth}{%
    \emph{Note:} \footnotesize Workload suffixes are target generator parameter $\mu_k/\sigma_k$.
    }
\end{table}

\section{Related Work}

\subsection{LLM Serving Systems} General inference optimizations include paged attention~\cite{kwon2023efficient}, prefix-aware KV reuse~\cite{zheng2024sglang}, and disaggregated prefill/decode~\cite{patel2024splitwise, zhong2024distserve}. 
However, these approaches generally assume a fixed routing configuration during serving, and optimize how to execute a given routing decision efficiently, rather than exposing routing top-$k$ itself as a runtime control operating point.

\subsection{MoE Architecture Adaptation and Fine-Grained Sparsity} 

Existing approaches adapt model computation along different axes. Compression techniques such as expert pruning~\cite{lu2024not} and quantization~\cite{frantar2022gptq, xiao2023smoothquant, ashkboos2024quarot} remove or approximate experts to obtain a smaller static model, while dense-to-MoE conversion~\cite{pei2025cmoe,zhu2024llama,qu2024llama} restructures dense models to introduce sparse expert activation. At runtime, DynMoE and Huang et al.~\cite{guo2024dynamic,huang2024harder} adapt the number of activated experts per token, whereas MoNE~\cite{cheng2025mixture} further explores finer-grained neuron-level conditional computation and EMoE~\cite{gu2025elastic} learns elastic top-$k$ routing through specialized training. Early-exit methods~\cite{schuster2022confident,elhoushi2024layerskip} adjust computation along the depth dimension, while D2MoE~\cite{wang2025d2moe} controls expert bit width and jointly optimizes I/O and compute scheduling. 
DualSparse-MoE~\cite{cai2025dualsparse} is close to our model-side design in that it also performs expert partitioning, but its primary goal is to coordinate tensor- and neuron-level sparsity for efficient MoE inference. In contrast, \modelname uses partitioning to create fine-grained intermediate routing budgets, then designs the scheduler and graph executor so that those budgets remain effective under request-level dynamic-$k$ serving.

\subsection{QoS-Aware Scheduling} 

QoS-aware scheduling has long been studied in latency-sensitive serving systems. Classical system works such as Minos~\cite{didona2019size} and Shenango~\cite{ousterhout2019shenango} reduce interference among heterogeneous requests through cost-aware resource management and fine-grained scheduling.
Scheduling is also widely explored in LLM serving. Representative examples include iteration-level scheduling in Orca~\cite{Orca}, throughput-latency-aware batching in Sarathi-Serve~\cite{agrawal2024taming}, multiplexing-aware scheduling in MuxServe~\cite{duan2024muxserve}, cross-instance rescheduling in Llumnix~\cite{sun2024llumnix}, and state-aware SLO scheduling in SOLA~\cite{hong2025sola}. 
These approaches do not explicitly schedule around an externally supplied MoE routing budget. Our scheduler consumes such targets and reduces cross-$k$ interference to avoid unnecessary over-serving.

\section{Conclusions}

We introduced \modelname, a model-system co-design that turns MoE routing top-$k$ from a fixed architectural setting into a practical serving-time control knob. \modelname combines a \emph{Model Refactoring Engine}, with training-free and optional gate-only fine-tuning paths, with a \emph{$k$-Aware Serving Runtime} comprising a \emph{$k$-Aware Runtime Scheduler} that preserves the computation savings of heterogeneous routing budgets and a \emph{Segmented Graph Executor} that retains CUDA Graph efficiency under heterogeneous routing budgets.
A current limitation of \modelname is that our evaluation is limited to the available two-GPU testbeds, while a two-GPU setup is the minimal non-trivial environment that successfully validates \modelname's core mechanism.

\bibliographystyle{IEEEtran}
\bibliography{sample-base}

\end{document}